# Illuminate: A novel approach for depression detection with explainable analysis and proactive therapy using prompt engineering.

Aryan Agrawal
aryanagrawal@ucsb.edu

## ABSTRACT

Traditional methods of depression detection on social media forums can classify whether a user is depressed, but they often lack the capacity for human-like explanations and interactions. This paper proposes a next-generation paradigm for depression detection and treatment strategies. This research employs three advanced Large Language Models (LLMs) - Generative Pre-trained Transformer 4 (GPT - 4) [1], Llama 2 chat [2], and Gemini [3], each fine-tuned using specially engineered prompts to effectively diagnose, explain, and suggest therapeutic interventions for depression. These prompts are designed to guide the models in analyzing textual data from clinical interviews and online forums, ensuring nuanced and context-aware responses. The study introduces a novel approach to prompt engineering, utilizing a few-shot prompting methodology for the Diagnosis and Explanation component. This technique is optimized to provide Diagnostic and Statistical Manual of Mental Disorders, Fifth Edition, (DSM-5) based analysis and explanation, enhancing the models ability to identify and articulate depressive symptoms accurately. For the Interaction and Learning aspect, the models engage in empathetic dialogue management, guided by resources from PsychDB [4] and a Cognitive Behavioral Therapy (CBT) Guide [5]. This facilitates meaningful interactions with individuals facing major depressive disorders, fostering a supportive and understanding environment. Furthermore, the research innovates in Case Conceptualization and Treatment, creating the Illuminate Database to guide the models in offering personalized therapy. This database is enriched with various CBT modules, encompassing case conceptualization, treatment planning, and therapeutic techniques. The models utilize this information to offer structured, actionable steps for addressing mental health issues. The quantitative analysis of the study highlights the effectiveness of these LLMs, demonstrated through metrics such as F1 scores, Precision, Recall, Cosine similarity, and Recall-Oriented Understudy for Gisting Evaluation (ROUGE) [6] across different test sets. This comprehensive approach, blending cutting-edge AI with established psychological methodologies, illuminates new possibilities in mental health care, showcasing the potential of LLMs in revolutionizing diagnosis and treatment strategies.

## I. INTRODUCTION

According to the World Health Organization (WHO), depression affects a staggering 300 million individuals globally, with an estimated 4.4% of the global population affected [7]. Despite this high prevalence, a significant portion of those experiencing depression remain undiagnosed or receive inadequate treatment, with misdiagnosis rates ranging between 30% to 50% [8].

The social stigma attached to mental health conditions, notably depression, acts as a substantial barrier to seeking necessary help. Studies indicate that stigma contributes to a significant delay in treatment-seeking behavior, with nearly 50% of individuals affected by depression refraining from seeking help due to stigma-related concerns [9]. Beyond immediate implications, depression carries substantial long-term effects, including an increased risk of comorbid health conditions like cardiovascular diseases, decreased quality of life, and substantial productivity loss. This condition's economic burden is staggering, accounting for an estimated $1 trillion in global costs annually [10].

Existing depression detection models yield binary classifications, categorizing individuals as either depressed or non-depressed, without offering accompanying explanations [11] [12]. These models lack transparency and fail to provide actionable insights, thereby presenting predictions without clear avenues for preventive actions. This limitation diminishes the persuasiveness of the classification outcomes for users and offers limited reference information for medical practitioners.

## II. ENGINEERING GOAL

The completion of this research aims to introduce "Illuminate", a novel approach for depression detection, explanation, personalized therapy, and prevention. Illuminate uses multiple data sources to build an explainable AI model that provides accurate personalized predictions and also offers transparent explanations for its decisions. Illuminate seamlessly integrates advanced Large Language Models (LLMs) into a mobile application, revolutionizing the way depression is diagnosed, explained, and addressed. By leveraging the capabilities of LLMs, including GPT-4, Llama-2-70b-chat, and Gemini, the application offers users a comprehensive mental health support system that combines human-like explanations and interactions with advanced AI-driven diagnosis and therapy. This research endeavors to bridge the gap between prediction and action, fostering a new paradigm for combating depression by leveraging AI models and prompting proactive interventions through the Illuminate mobile application.

## III. METHODS

The overall architecture of this research is illustrated in FIG. 1 and FIG. 2. The first part focuses on implementing Machine Learning models, while the second part delves into the use of Large Language Models for depression detection, explanation, and therapy.

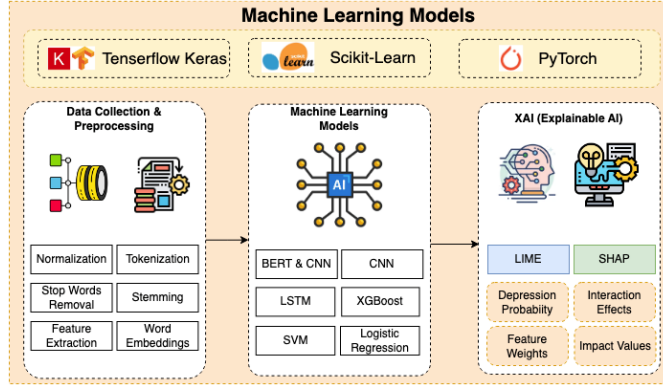

**FIG. 1.** Architecture for Machine Learning Models

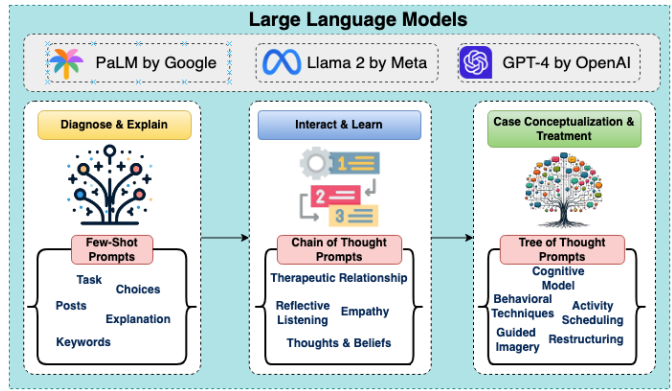

**FIG. 2.** Architecture for Large Learning Models

### A. Datasets

DAIC-WOZ Depression Database: This dataset is part of a larger corpus, the Distress Analysis Interview Corpus (DAIC) [13] which comprises clinical interviews aimed at facilitating the diagnosis of mental health issues such as anxiety, depression, and post-traumatic stress disorder. These interviews are part of a larger initiative to develop a computer agent capable of conducting interviews to detect verbal and nonverbal signs of mental illness [14] [15]. The data gathered includes audio and video recordings, along with comprehensive questionnaire responses. The data has been transcribed and annotated for various verbal and nonverbal attributes. The dataset contains interviews of 189 users, 130 of those are non-depressed and 59 are depressed. This research utilizes the transcript files available in the dataset.

Reddit Mental Health Dataset (RMHD): Reddit mental health dataset contains posts between 2018 and 2020 from 17 mental health and 11 non-mental health subreddits [16]. Subreddits dedicated to mental health topics such as depression, anxiety, and related concerns serve as online communities where individuals share experiences offer support, and exchange information regarding mental health challenges. These subreddits provide a platform for discussions, seeking advice, sharing coping strategies, and connecting with others who may be going through similar experiences. This research paper utilizes posts from r/depression, r/suicidewatch, r/mentalhealth, r/legaladvice, and r/teaching. To use this unlabeled dataset, I performed semi-supervised learning with pseudo-labeling [17] using the Huggingface model MentalRoBERTa. [18]

The combined RMHD and DAIC-WOZ dataset was split into training, validation, and testing data using a 60-20-20 split. Table 1. shows the number for each of the classes in the different datasets.

| Datasets | Labels | | Total Posts |
|---|---|---|---|
| | Depressed (1) | Not Depressed (0) | |
| Training Set | 109,768 | 107,230 | 216,998 |
| Validation Set | 13,628 | 13,497 | 27,125 |
| Testing Set | 13,619 | 13,506 | 27,125 |

**TABLE 1.** Dataset Distribution for Training, Validation, and Testing

### B. Data Preprocessing

To prepare the dataset for deep learning models targeting depression detection, I first normalized the text by converting it to lowercase, removing punctuations, and eliminating noise such as non-ASCII characters. I then tokenized the text, breaking it down into individual words, and streamlined it by removing stop words. This process reduced unnecessary data complexity, allowing the model to concentrate on meaningful content. I applied stemming next to condense words to their base forms, ensuring consistency in the dataset. Finally, I transformed the text into numerical vectors through feature extraction and word embeddings, capturing the contextual relationships between words. These preprocessing steps were crucial for enhancing the model's ability to interpret and analyze the textual data accurately, setting a strong foundation for efficient and effective depression detection.

### C. Models Architecture and Implementation

I implemented six machine learning models primarily using *scikit-learn* and *keras*.

**C.1 Machine Learning Models**
The traditional machine learning models included a Support Vector Machine (SVM) [19] with a Radial Basis Function (RBF) kernel, XGBoost [20], and Logistic Regression [21]. Hyperparameters were optimized using the $10-fold$ k-cross validation. Key parameters optimized for the SVM included the penalty parameter C and the kernel coefficient . For XGBoost, parameters such as learning rate , max depth of trees maxdepth, and the number of trees nestimators were fine-tuned. Logistic Regression utilized a L2 regularization parameter and a liblinear solver for predicting the depression.

**C.2 Deep Learning Models**
For the deep learning models, I utilized *keras* library to develop a Convolutional Neural Network (CNN) model, a hybrid Bidirectional Encoder Representations from Transformers (BERT) - CNN model [22], and a Long Short-Term Memory (LSTM) model [23].

**C.2.1 CNN Model Architecture**
The CNN model architecture uses the pre-trained Word2Vec embeddings from the Google news dataset. An embedding matrix is created mapping each word in the model's vocabulary to its corresponding 300-dimensional vector from $Word2Vec$ [24]. The first convolutional layer utilizes $24\ filters$, with a $kernel-size = 10$, and employs the $relu$ (Rectified Linear Unit) activation function. This layer is pivotal in extracting initial features from the input data. Following this, the second convolutional layer comprises $8\ filters$, each with a smaller $kernel-size = 5$, also using the $relu$ activation function. This layer serves to further refine the features extracted by the preceding layer. Max pooling layers follow each convolutional layer, reducing the dimensionality and thus the computational complexity, while retaining important features. A $Flatten$ layer is used to convert the 2D feature maps into a 1D vector, enabling the transition to dense layers. A dropout layer with a rate of 30% is included to reduce overfitting.

Finally, the model employs a $Dense$ output layer with two neurons and a $sigmoid$ activation function, suitable for binary classification. The model is compiled with the $adam$ optimizer and $binary\ cross-entropy$ loss, which is effective for binary classification tasks in neural networks.

**C.2.2 BERT-CNN Model**
The BERT-CNN model is developed using embeddings generated from pre-trained Bidirectional Encoder Representations from Transformers (BERT) Sentence Transformer $nq-distilbert-base-v1$. The CNN model starts by adding the $Reshape$ layer to convert 768-dimensional BERT embeddings with the following Conv1D layer. The rest of the CNN architecture is similar to the standalone CNN model described in the CNN model architecture.

**C.2.3 Long Short-Term Memory (LSTM) Model**
The LSTM model starts with an embedding layer, configured for a vocabulary of 20,000 words, transforming words into 128-dimensional vectors. This setup is designed to process up to 800 words of input sequences. Following the embedding layer is an LSTM layer with 128 units, featuring both $dropout$ and $recurrent-dropout$ set at 20%, which helps in mitigating overfitting by randomly omitting a portion of the inputs and recurrent connections during training. An additional $Dropout$ layer with a rate of 50% is introduced after the $LSTM$ layer for further regularization, enhancing the model's ability to generalize to unseen data. The model concludes with a $Dense$ layer, employing a $sigmoid$ activation function, making it suitable for binary classification tasks. Compiled with $adam$ optimizer and $binary\ cross-entropy$ loss, the model is trained and validated over 20 $epochs$ with a $batch-size$ of 64, aiming to capture temporal dependencies and contextual nuances in text data effectively.

**D. Models Performance**
Performance of all 6 models is measured using Accuracy, Precision, Recall, and F1-Score. Results are listed in Table 2.

**D.1 Accuracy**
Accuracy is the simplest and most intuitive performance metric. It is the ratio of correctly predicted observations to the total observations. It is defined as:
$$Accuracy = \frac{Number\ of\ correct\ predictions}{Total\ number\ of\ predictions}$$

**D.2 Precision**
Precision is the ratio of correct positive predictions to the total predicted positive predictions. It is the measure of model's exactness. It is defined as:
$$Precision = \frac{True\ Positives}{True\ Positives + False\ Positives}$$

**D.3 Recall**
Recall is the ratio of correct positive predictions to the actual positives in the dataset. It is defined as:
$$Accuracy = \frac{True\ Positives}{True\ Positives + False\ Negatives}$$

**D.4 F1-Score**
F1-Score is the weighted average of Precision and Recall. It is a better measure than accuracy for binary classification. It is defined as:
$$F1-Score = 2 * \frac{Precision * Recall}{Precision + Recall}$$

| Model | Accuracy | Precision | Recall | F1-Score |
|---|---|---|---|---|
| BERT-CNN | 0.94 | 0.91 | 0.95 | 0.94 |
| LSTM | 0.85 | 0.76 | 0.74 | 0.76 |
| CNN | 0.85 | 0.72 | 0.87 | 0.79 |
| SVM | 0.52 | 0.48 | 0.45 | 0.43 |
| XGBOOST | 0.72 | 0.67 | 0.71 | 0.70 |
| LOGISTIC REGRESSION | 0.63 | 0.52 | 0.48 | 0.53 |

**TABLE 2.** Model Performance measured using Accuracy, Precision, Recall, and F1-Score for the Machine Learning Models.

### E. Explainable AI

To enhance the credibility of our model, we employ rigorous validation techniques, including multiple cross-validations and test set validations. These methods provide a comprehensive assessment of the model's performance on unfamiliar data sets, ensuring its robustness and reliability. However, these validations fall short of offering insights into the underlying reasons for specific predictions made by the model. Local Interpretable Model-agnostic Explanations (LIME) explains a prediction so that black-box models can be represented in a human-interpretable way [25]. I have used the BERT-CNN model to interpret the test results. It provides probability and the tokens which contributed to the probability. The results of 4 predictions are listed in TABLE 3.

## IV. LARGE LANGUAGE MODELS AND PROMPT ENGINEERING

Large Language Models (LLMs) are advanced AI systems trained on vast amounts of text data, capable of understanding and generating human-like language responses across various topics. Prompt engineering [26] is a relatively new discipline of skillfully crafting and optimizing prompts to effectively guide these models for a wide variety of applications and maximize the potential of their language generation capabilities. High-level architecture for the prompts engineering and guiding the Large Language models is shown in FIG. 3.

I have implemented three distinct prompt managers targeting the Diagnosis, Interaction, and Prevention domains. The first prompt for Diagnosis and Explanation is developed using a few-shot prompting methodology [27]. The prompt is optimized to provide the DSM-5 based analysis and explanation. For Interaction and Dialogue management, a $Chain\ of\ Thought\ (CoT)$ prompting [28] framework is used. The third prompt manager, focused on prevention and therapy, integrates a $Tree\ of\ Thought\ (ToT)$ prompting approach [29]. This approach is pivotal in administering personalized therapy, utilizing iterative feedback from the other 2 models.

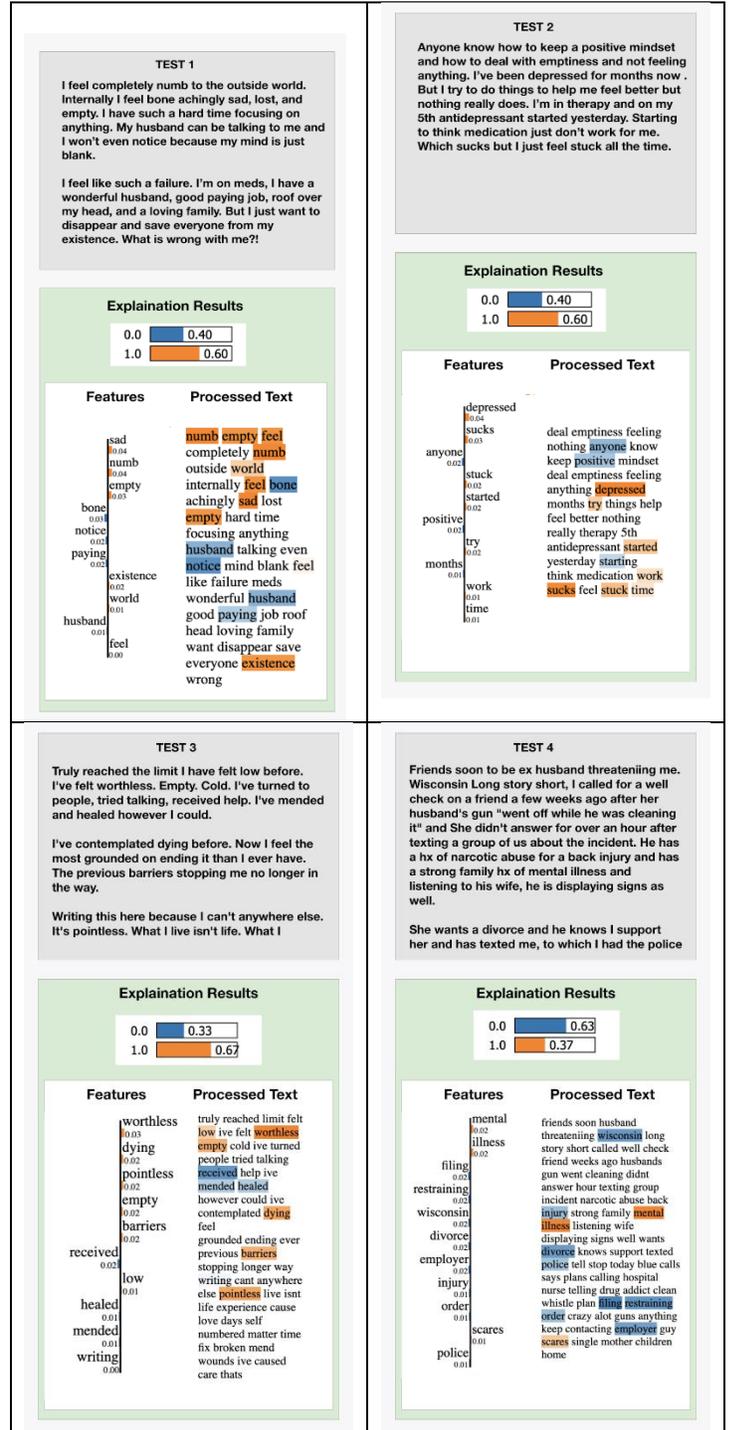

**TABLE 3.** Explanation and Probability of Depression using LIME. The first 3 tests are with the true label "Depressed", and 4th test is with the true label "Not Depressed."

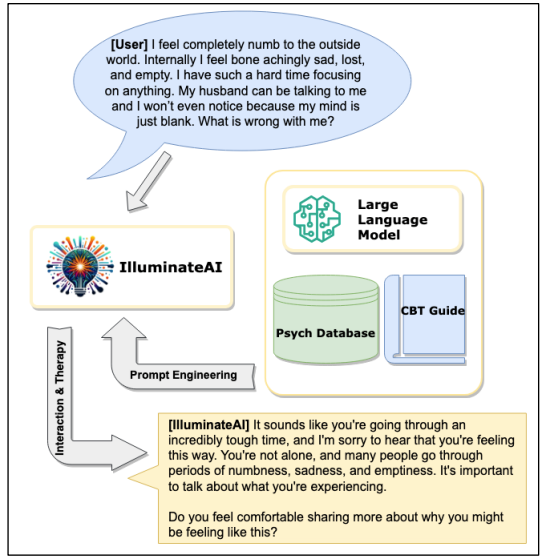

**FIG. 3.** High level architecture to fine-tune large language models for interaction, dialogue, and treatment.

## A. Diagnose & Explain

The first prompt for Diagnosis and Explanation is developed using a few-shot prompting methodology. Including examples in a prompt is crucial as they guide the model in understanding and accurately performing the task, set clear expectations for the desired output, and act as a mini-training set to enhance the model's response accuracy. The prompt is shown in TABLE 4.

| Diagnose & Explain Prompt |
|---|
| You are an assistant trained in non-clinical text analysis based on the Diagnostic and Statistical Manual of Mental Disorders, Fifth Edition (DSM-5). Your role involves examining written posts, identifying potential signs of depression, and providing explanations based on DSM-5 criteria.<br>**Structure of Each Example:**<br>**Task:** Read and analyze posts to determine the presence of depressive symptoms as defined by DSM-5.<br>**Choices:** [A. Depressed B. Not Depressed]<br>**Posts:** [Text from the post]<br>**Answer:** [A/B based on analysis]<br>**Explanation:** Provide a concise and scientific rationale for your decision, citing specific DSM-5 criteria.<br>**Keywords:** List keywords or phrases from the post that are indicative of depression. |

**TABLE 4.** Prompt for Diagnosis and Explanation using DSM-5.

## B. Interact and Learn

This prompt is based on resources from PsychDB (Cognitive Behavioural Therapy (CBT), 2024) and a Cognitive Behavioral therapy guide from Washington University. It guides the model in a non-clinical, experimental setting to interact empathetically with individuals facing mental health challenges. It follows a Chain of Thought (CoT) approach, as shown in FIG. 4, starting with understanding the emotions and context of the person's posts and then relating these insights to the provided explanation of their mental state.

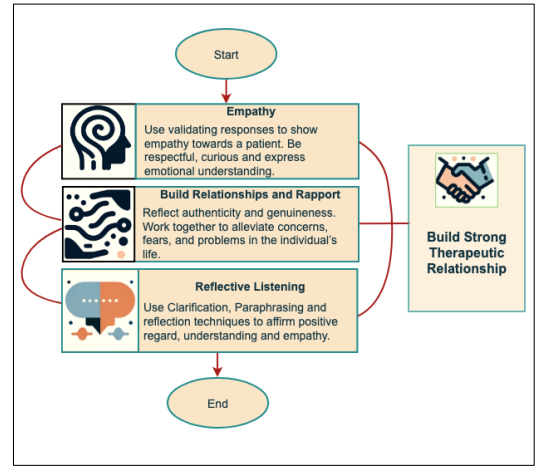

**FIG. 4.** Prompt Design for Dialogue and Interaction

The model formulates questions to delve deeper into the individual's feelings and thoughts, ensuring clarity and a deeper understanding of their situation, all while maintaining a supportive and empathetic tone throughout the conversation. A few shot examples are listed in Table 5 for Clarification, and Table 6 for Paraphrasing and Reflection.

| Clarification |
|---|
| **[User]** I just do not feel like trying anymore.<br>**[Illuminate AI]** Tell me more about what you mean.<br>**[User]** I just feel like giving up.<br>**[Illuminate AI]** Do you mean giving up on your goal to complete college; or are you referring to something different, like giving up on life and possibly harming yourself?<br>**[User]** I am not referring to suicide if that is what you mean, but I am feeling really depressed. Each day seems like such a struggle, and I often just feel like staying in bed. When I said "give up," I guess I was referring to not wanting to face all the struggles I face in life - my school work, financial problems, relationship problems, etc. |

**TABLE 5.** A few shot examples for clarification via dialogues.

| Paraphrasing & Reflection |
|---|
| **[User]** Since my fiancé's death, I feel like every day is a struggle, and I often question whether my life will ever get better. I just miss him so much that I think about him constantly. I don't know what to do, but the pain is getting to be too much.<br>**[Illuminate AI]** You are really struggling to feel better, and much of your pain comes from the grief and loss you feel from losing your fiancé. You may even be questioning whether or not this pain will subside because it is getting unmanageable.<br>**[User]** Yes, that is correct. I do not want you to think that I think only about the pain of losing him. The pain I feel comes from my intense feelings of loss, but this pain is also because |

> I miss all the things he meant to me, and the joy he brought to my life. I am really struggling because I do not want to let go of him, but holding on hurts so much.

**TABLE 6.** A few shot examples for paraphrasing and reflection via dialogues.

## C. Case Conceptualization and Treatment

I created the Illuminate Database to guide the model towards conceptualization and treatment. The guide contains detailed modules covering various aspects of CBT, including case conceptualization, treatment planning, identifying and challenging maladaptive thoughts, behavioral activation, problem-solving, and relaxation techniques. To create a structured database, I extracted key elements from each module, focusing on the specific CBT techniques, goals, and approaches recommended for different mental health issues. It also includes extracting detailed steps for each technique, the objectives of each module, and the specific approaches to case conceptualization and treatment planning.

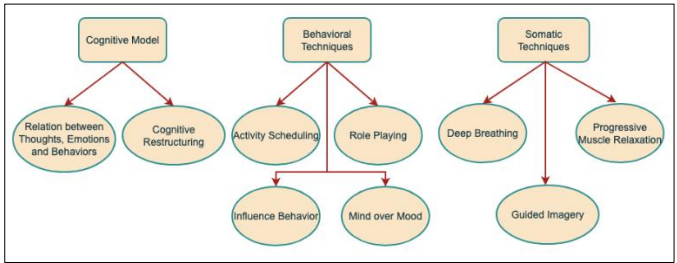

**FIG. 5.** Prompt Design for Case Conceptualization and Treatment.

Each module serves as a node in the ToT (Tree of Thought) prompting technique as shown in FIG. 5. A few shot examples for Cognitive restructuring are listed in TABLE 7, and for Behavioral Activation in TABLE 8.

**Node - Cognitive Restructuring**

**[Objective]** To help clients identify and challenge maladaptive thoughts.
**[Techniques]** Thought Recording, Evidence Gathering, Thought Challenging.
**[Application]** Use in cases of anxiety and depression where distorted thinking patterns are evident.
**[Prompt Example]** Identify a recent situation where you felt unusually anxious or depressed. What were your thoughts at that time? How did these thoughts affect your feelings and behavior?
**[Key Steps]** Identifying automatic thoughts, examining the evidence for and against these thoughts, and developing more balanced thoughts.

**TABLE 7.** A few shot examples for cognitive restructuring.

**Node - Behavioral Activation**

**[Objective]** To increase engagement in positively reinforcing activities.
**[Techniques]** Activity Scheduling, Pleasure Predicting.
**[Application]** Particularly useful in cases of depression where inactivity or avoidance behaviors are present.
**[Prompt Example]** List activities you used to enjoy or might enjoy. Plan to engage in one of these activities this week. How do you expect to feel during and after the activity?
**[Key Steps]** Identifying low-mood patterns, scheduling enjoyable activities, and monitoring mood changes associated with activities.

**TABLE 8.** A few shot examples of behavioral activation.

## V. RESULTS

In this study, I utilized three large language models to conduct the experiment: GPT-4, Llama-2-70b-chat, and Gemini. These models represent the forefront of AI capabilities in language understanding and generation. All the models were fine-tuned by employing three distinct prompt managers tailored to the domains of Diagnosis, Interaction, Therapy, and Prevention for mental health. The Diagnose & Explain Prompt Manager utilized a few-shot prompting methodology, emphasizing the incorporation of examples to guide the models. These prompts were designed for models to operate as non-clinical text analysts, identifying potential signs of depression based on DSM-5 criteria. Each example involved analyzing posts, determining the presence of depressive symptoms, and providing concise explanations, citing specific DSM-5 criteria and indicative keywords. The Interaction prompt employing a $Chain\ of\ Thought\ (CoT)$ framework fostered empathetic interactions and allowed the model to learn more about the user. The Prevention & Therapy Prompt Manager adopted a $Tree\ of\ Thought\ (ToT)$ technique and utilized an Illuminate Database to guide models towards case conceptualization and treatment.

### A. Quantitative Evaluation

The performance evaluation of these LLMs consisted of several quantitative measures:

**F1-Score:** The performance of the 3 models GPT-4, Gemini Pro, and LLama-2 in response to the 'Diagnose & Explain' prompt was evaluated using the F1-Score. The F1-Scores were measured with varying numbers of few-shot prompts (1-shot, 2-shot, 3-shot, and 4-shot prompts) for each model. GPT-4 achieved the highest F1-Score of 0.95 when using 4-shot prompts, outperforming Gemini Pro and LLama-2, which achieved F1-Scores of 0.83 and 0.89, respectively, with the same number of prompts. It was also noticed that performance does not vary much between 3 and 4-shot prompts and results are shown in FIG. 6.

**Cosine Similarity:** For the 'Interaction and Learning' and 'Case Conceptualization & Treatment' prompts, due to the collaborative nature, the results are combined, and performance is measured using Cosine Similarity. This score quantifies the similarity between two texts by representing them as vectors in a space and calculating the cosine of the angle between these vectors. This assessment was conducted on a dataset comprising 300 posts from 100 distinct users. I analyzed the results at various data sizes, including 25%, 50%, 75%, and 100% of the test data and results are shown in FIG. 7.

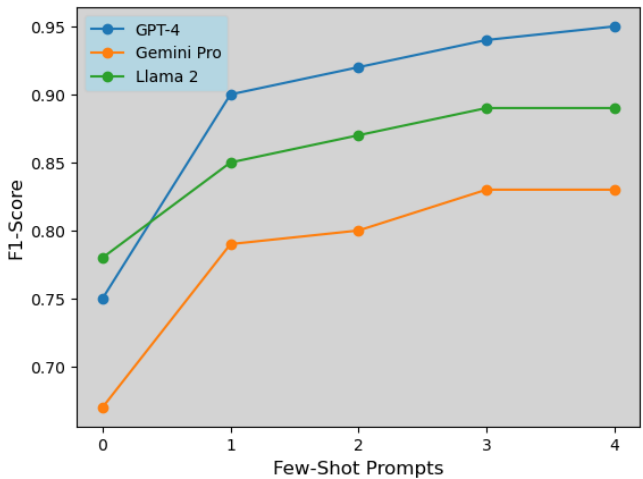

**FIG. 6.** F1-Score for fine-tuned models using 1, 2, 3, and 4 shot prompts.

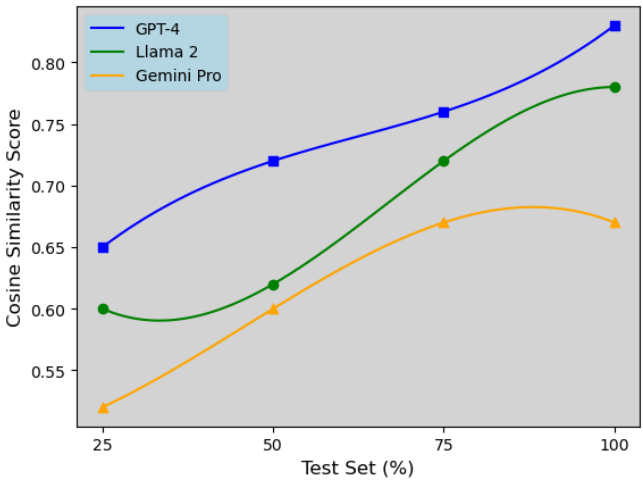

**FIG. 7.** Cosine Similarity Score for fine-tuned models using 25%, 50%, 75% and 100% of the Test Set.

**Recall-Oriented Understudy for Gisting Evaluation (ROUGE) metric:** This score measures the quality of text summaries by comparing them with reference summaries, focusing on the overlap of n-grams, word sequences, and sentence sequences to assess the similarity and coverage. This assessment was also conducted on a dataset comprising 300 posts from 100 distinct users. Results for F1-Score, Precision and Recall are listed in Table 9.

| Model | Precision | Recall | F1-Score |
|---|---|---|---|
| GPT-4 | 0.78 | 0.75 | 0.74 |
| Gemini Pro | 0.66 | 0.64 | 0.66 |
| Llama 2 | 0.72 | 0.67 | 0.69 |

**TABLE 9.** Precision, Recall and F1-Score for fine-tuned models using ROUGE metric.

## VI. MOBILE APPLICATION

The development of the IlluminateAI mobile application represents a significant leap forward in deploying AI for mental health support directly into user's hands. Built upon a fine-tuned GPT-4 model, IlluminateAI is designed to provide immediate, personalized assistance through its innovative features. Using the **Diagnose** feature, the model predicts, provides analysis, an explanation behind the analysis, and keywords. Sample results of the diagnose feature for 4 posts from different users are listed in FIG. 8 below. The app's **Chat** feature is a combination of "Interaction and Learning" and "Case Conceptualization & Treatment" prompts, which facilitate a dialogue-based interaction between the user and AI. These features are illustrated in the application's screenshots in FIG. 9, demonstrating the intuitive and engaging user interface designed to foster an environment of empathy and support. Through these dialogues, IlluminateAI adapts to each user's unique context, providing tailored advice, exercises, and coping strategies derived from cognitive-behavioral therapy (CBT) principles and PsychDB, which are integrated into the models while engineering the prompts.

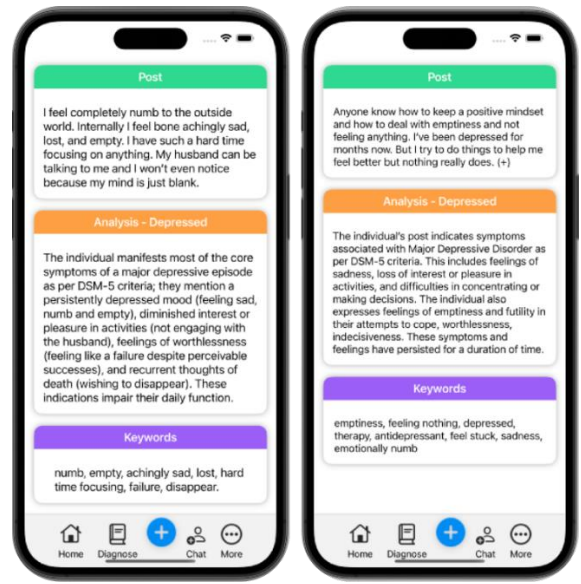

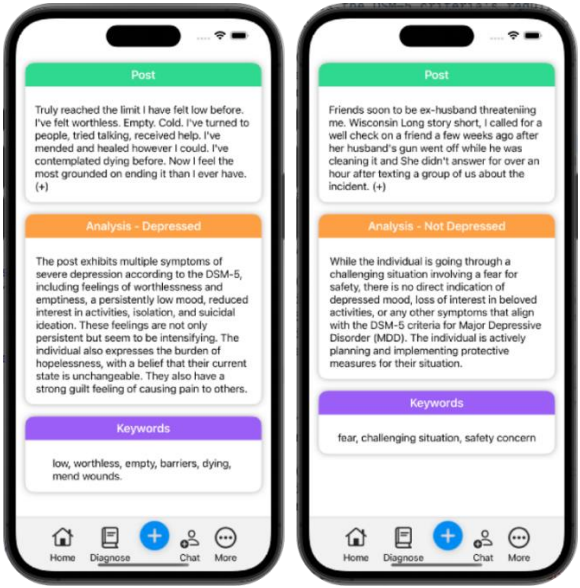

**FIG. 8.** Analysis of test posts with Explanation and Keywords using the fine-tuned GPT-4 model.

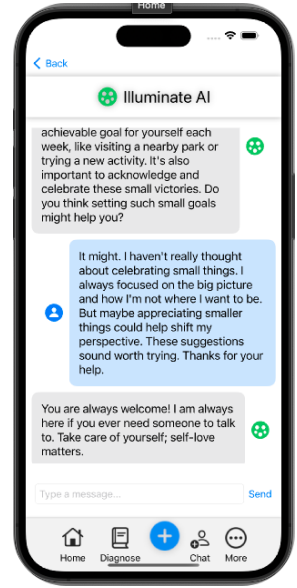

**FIG. 9.** Sample of Interactive Dialogue between the user and Illuminate AI.

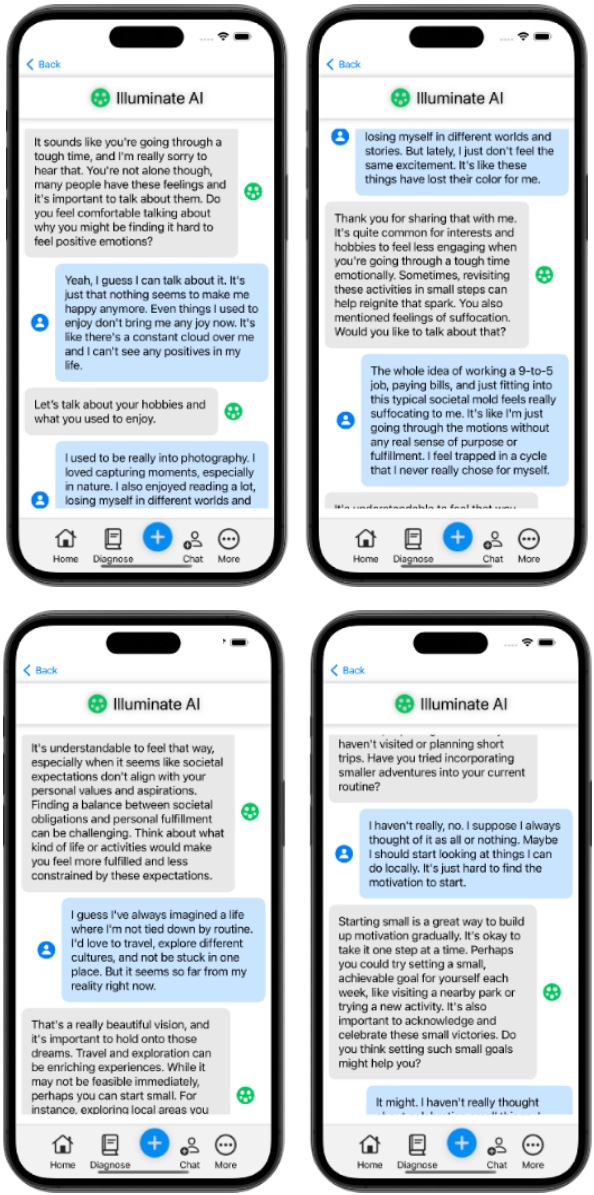

## VII. CONCLUSION

This research provides an innovative framework for depression detection and intervention, leveraging the capabilities of prompt engineering and advanced language models. Through the development of IlluminateAI, I have demonstrated a novel method that marries AI's analytical precision with the nuanced understanding required for empathetic mental health support. My findings indicate significant advancements in user engagement and diagnostic accuracy, suggesting a promising avenue for reducing the stigma and accessibility barriers associated with mental health care. By integrating AI with empathetic, human-like interactions, IlluminateAI not only enhances the immediacy and personalization of support for individuals with depression but also offers a scalable solution to meet widespread mental health needs effectively. The study acknowledges its limitations to major depressive disorders only. Future research should focus on refining these AI models to encompass a wider spectrum of mental health conditions and integrating user feedback to further personalize and improve the efficacy of mental health interventions. This research sets the stage for a transformative approach to mental health care, emphasizing the potential of AI to complement traditional therapeutic models and contribute to a more accessible, responsive, and comprehensive mental health ecosystem.

## REFERENCES

[1] GPT-4. (n.d.). https://openai.com/gpt-4

[2] meta-llama/Llama-2-70b-chat-hf · Hugging Face. (n.d.). https://huggingface.co/meta-llama/Llama-2-70b-chat-hf